\documentclass[10pt,conference,a4paper]{IEEEtran}
\usepackage{graphicx}
\usepackage{amsmath}
\usepackage{url}
\usepackage{multirow}
\usepackage{color}

\IEEEoverridecommandlockouts

\begin{document}
\title{CNN+RNN Depth and Skeleton based Dynamic Hand Gesture Recognition}

\author{
\IEEEauthorblockN{Kenneth Lai and Svetlana N. Yanushkevich}
\IEEEauthorblockA{Biometric Technologies Laboratory, Department of Electrical and Computer Engineering\\
University of Calgary, Alberta, Canada\\
Email:kelai@ucalgary.ca, syanshk@ucalgary.ca}}

\maketitle
\IEEEpubid{\begin{minipage}{\textwidth}\ \\[70pt]
		\textbf{\footnotesize{{\fontfamily{ptm}\selectfont Digital Object Identifier 10.1109/ICPR.2018.8545718 \\978-1-5386-3788-3/18/\$31.00 \copyright 2018 IEEE}}}
\end{minipage}}
\begin{abstract}
Human activity and gesture recognition is an important component of rapidly growing domain of ambient intelligence, in particular in assisting living and smart homes.  In this paper, we propose to combine the power of two deep learning techniques, the convolutional neural networks (CNN) and the recurrent neural networks (RNN),  for automated hand gesture recognition  using both depth and skeleton data. Each of these types of data can be used separately to train neural networks to recognize hand gestures. While RNN were reported previously to perform well in recognition of  sequences of movement for each skeleton joint given the skeleton information only, this study aims at utilizing depth data and apply CNN to extract important spatial information from the depth images.  Together, the tandem CNN+RNN is capable of recognizing a sequence of gestures more accurately. As well, various types of fusion are studied to combine both the skeleton and depth information in order to extract temporal-spatial information. An overall accuracy of 85.46\% is achieved on the dynamic hand gesture-14/28 dataset.
\end{abstract}
\IEEEpeerreviewmaketitle

\section{Introduction}
This paper concerns with biometric-based ambient computational intelligence techniques.  Specifically, we focus on human activity and gesture recognition in the context of ambient  monitoring in smart homes, assisting living or healthcare facilities.  In a smart home, sensors can be programmed to learn about a resident's normal daily routines which can then be used for performing automated ambient health monitoring and assessment \cite{dawadi2013automated}.  For example, Pavel et al. \cite{pavel2006mobility} suggested that there is a relationship between mobility patterns and cognitive ability.  The theory was examined by observing the changes in mobility and found evidence to support the relationship between mobility and cognitive ability. In the current aging population, ``the challenges of maintaining mobility and cognitive function make it increasingly difficult to remain living alone therefore forcing many people to seek residence in clinical institutions'' \cite{arcelus2007integration}.

Lee and Dey \cite{lee2010embedded} designed an embedded sensing system to determine if the resident gains more awareness about their functional abilities when given information regarding their movements. The ability to perform automated assessment of task quality and cognitive health has greatly improved accuracy \cite{cook2010learning,kim2010human}. These techniques indicate that specific information can be extracted from a sensor and used in labelling the performed activity. For example, some activities such as washing dishes, taking medicine, and using the phone are characterized by the interaction with unique objects. 

The main objective of this paper is to implement a framework for activity recognition, including gesture recognition.  Traditional activity recognition uses mainly RGB images for analysis.  Current methods incorporate different types of information including depth, infrared, and time \cite{koppula2013learning}.  The proposed method focuses on creating a framework that uses both depth and skeleton information in the task of hand gesture/activity recognition.  To prove such point, we have selected the specific task of recognizing dynamic hand gestures using depth and and skeleton information. We apply the state-of-the-art deep learning techniques such as convolutional neural network (CNN) and recurrent neural networks (RNN).

The paper is structured as follows: related works are introduced in Section \ref{sec:relatedworks}, framework of the proposed method in Section \ref{sec:framework}, design of experiments and experimental results in Section \ref{sec:experiments}, and the conclusions in Section \ref{sec:conclusions}.

\section{Related Works}\label{sec:relatedworks}
Activity and gesture recognition is an actively researched domain, especially in the light of recent development of new and advanced types of sensors that collect multiple, more precise data.  Different spectra of data have been examined in the task of recognizing gestures and activities, including color, depth, GPS, acceleration, infrared, etc.

Prior to popularization of inexpensive depth sensors, it was mainly color (pixel intensity), or RGB, data available for gesture recognition.  One of the most common method was to wear a color markers that indicate  different regions of the hand.  Iwai et al. used a colored glove in combination with a decision tree to perform gesture recognition \cite{iwai1996gesture}.  Another color-based approach  by Bretzner et al. utilized multi-scale color features \cite{bretzner2002hand}.

With the development of Microsoft Kinect, LeapMotion, and Intel RealSense, depth and 3D information is now more easily acquired for analysis.  An approach based on extracted depth features that allow to create depth silhouette is proposed in \cite{munoz2008depth}.  Another method that uses depth images  to train a multi-layered random forest is suggested in \cite{kuznetsova2013real}.

Motivated by the relationship between hand gestures and sign language recognition, \cite{pugeault2011spelling} proposed a method of recognizing hand shapes using random forests applied to both depth and color images.  Another approach uses depth and color images and various types of spatiotemporal descriptors,  combined with different kernel choices for support vector machine to find the optimal recognition combination \cite{ohn2014hand}.

In  recent years, deep learning techniques have revolutionized the pattern recognition in general.  A 3D CNN \cite{molchanov2015hand}  combines spatiotemporal data augmentation in order to perform gesture recognition on both depth and color images. It has achieved a 77.5\% classification rate on the VIVA challenge dataset.  A method proposed by Nagi et al. \cite{nagi2011max}, suggest the use of ``state-of-the-art big and deep neural network combining convolution and max-pooling'' for feature extraction and classification.  In \cite{Chen2017}, RNN is used to model the temporal information in a sequence of skeleton joint movement to perform gesture recognition.  Similarly, \cite{Nunez2018} combines CNNs with long short term memory (LSTM) to recognize dynamic hand gesture using only skeleton-based information.

\section{Framework}\label{sec:framework}

\begin{figure}[!ht]
\begin{center}
\includegraphics[width=0.34\textwidth]{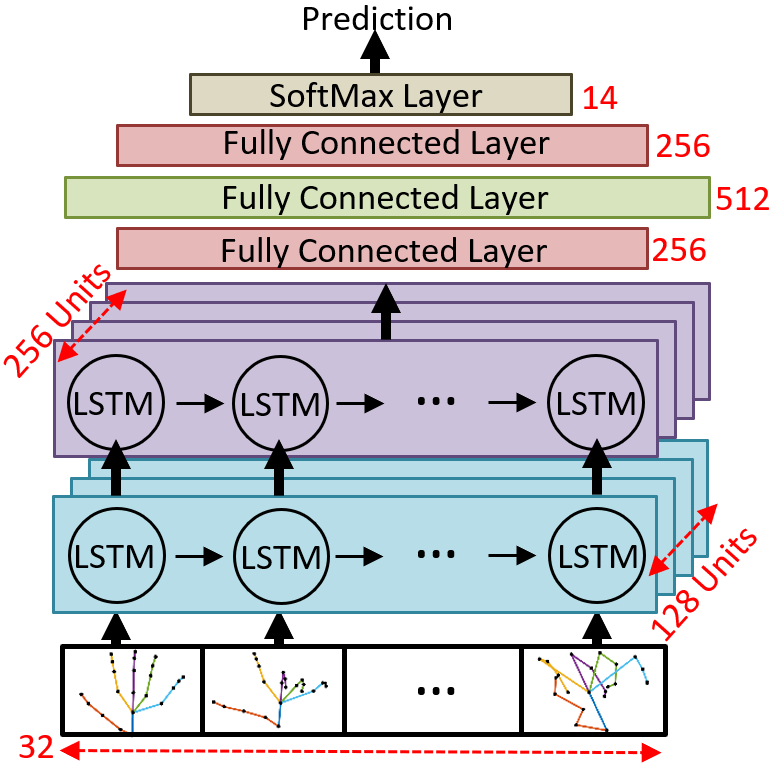} 
\caption{The structure of the skeleton-based LSTM network.}
\label{fig:lstm}
 \end{center}
\end{figure}

\begin{figure*}[!htb]
\begin{center}
\includegraphics[width=0.8\textwidth]{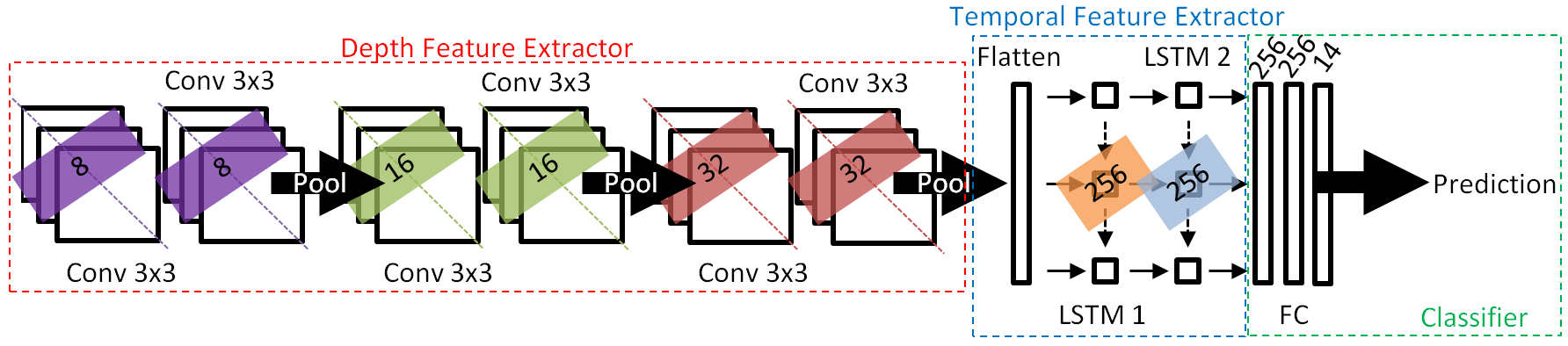} 
\caption{The structure of the depth-based CNN+LSTM network.}
\label{fig:cnn+lstm}
 \end{center}
\end{figure*}

This section describes the proposed approach of using both depth and skeleton points in recognizing a hand gesture. The system consists of two main components: a depth-based CNN+RNN (Fig. \ref{fig:cnn+lstm}), and a skeleton-based RNN (Fig. \ref{fig:lstm})


The first component of the system, CNN, is designed to extract features from depth image of a subject performing a hand gesture.  However, since the processing of dynamic gesture recognition is dependent on an ordered sequence of images, a RNN is proposed to supplement the CNN to extract temporal patterns.  The overall architecture of the CNN and the Long Short Term Memory (LSTM) network is shown in Fig. \ref{fig:cnn+lstm}.  The  structure consists of three components: the depth based feature extraction through CNN, the time series processing through RNN, and classification using a multilayer perceptron (MLP).  The first component,  CNN, includes six 3x3 convolutional layers with a 2x2 max pooling layer between every other convolutional layer. The second component,  RNN, includes two LSTM layers, each consisting of 256 LSTM units. The final component,  MLP, contains three fully connected (FC) layers consisting of 256, 256, and 14 units, respectively.

The second component of the system includes RNN which extracts temporal patterns from the movements of skeleton points within a sequence. The  RNN structure is shown in Fig. \ref{fig:lstm} and is similar to the one described in the first component, except for an additional FC layer and several LSTM units within each layer.  

Overall, the proposed framework consists of two networks components that uses skeleton and depth information for gesture recognition independently.  Since each network is capable of predicting a selected gesture based on the selected type of information, a fusion of both network is expected to yield a higher performance that is capable of selecting the best features from both the skeleton, the depth-based spatial information and the inherent temporal patterns between a collection of frames.  

\begin{figure}[!ht]
\begin{center}
\includegraphics[width=0.34\textwidth]{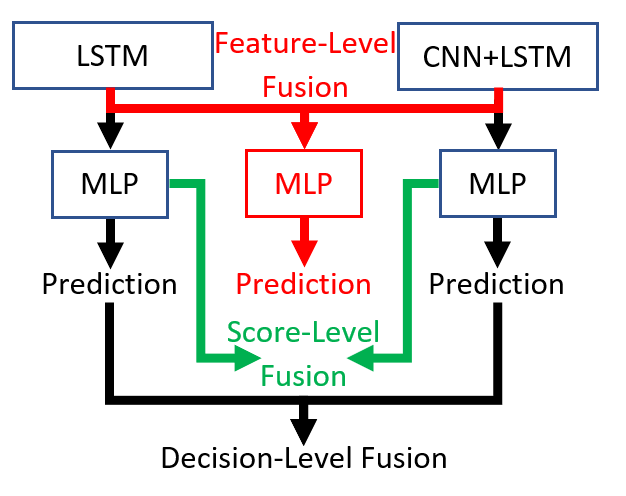} 
\caption{The three levels of fusion for the LSTM and CNN+LSTM networks.}
\label{fig:fusion}
 \end{center}
\end{figure}

 There are different ways to perform fusion as illustrated in Fig. \ref{fig:fusion}. The three main techniques considered in this study include the feature-level fusion, the score-level fusion, and the decision-level fusion.

The feature-level fusion can be performed at any layer before the MLP (which consists of the fully-connected, soft-max, and classification layers).  In general, the convolution layers in the CNN, and the LSTM layers in the RNN, are the portion of the network designed to extract features from the input data.  A feature-level fusion can be performed by a fusion of the results after the input data has passed through a series of convolution/LSTM layers.  Following the fusion, a classifier such as MLP or support vector machine can be attached in order to create a feature-level fusion network.  

Similarly to the feature-level fusion, the score-level fusion can be performed after or between the fully-connected and soft-max layers. In this process, we  assume that features have been successfully extracted from the input data and  passed through a selected classifier resulting in some score or probability.  Since the score is heavily correlated to the network's prediction, the combination of multiple scores from different networks will provide a more reliable prediction.

The decision-level fusion is comparable to a score-level fusion except that the fusion is performed after the network's prediction. This type of fusion is based entirely on the network's predicted output and is not associated with the score/probability used for the decision. A network's predicted output is defined based on the probability/score output from the soft-max layer.  The two most general method for a network to generate a prediction is either based on a decision threshold or the ranking order.  The decision threshold method is based on selecting an arbitrary value for each network.  Any scores above the selected value is considered accepted otherwise rejected. As for the ranking order method, all the scores are grouped and ordered such that the higher the list the more likely acceptance.  A general usage is the rank-1 recognition which only considers the top prediction as the network's predicted output.

In this paper, we have explored both the feature-level and score-level fusion but excluded the decision-level fusion.  Decision-level fusion is not examined in this paper because it is correlated to the score-level fusion, and there are only two network decisions to be combined.  

 In addition to the two level of fusion, there are various types of fusion of which we consider concatenation, averaging, and maximum.  Fusion using concatenation was performed in this study at the feature-level because it generates a new set of features that considers both the extracted depth and skeleton information.  Fusion using averaging and maximum is only applied to the score-level fusion because each score represents a network's strength in prediction.  The averaging method is expected to generate a more reliable score, since it relies on two types of information and networks. Finally, the maximum technique shall place more emphasis on each network's ability to predict specific gestures.  

\section{Experiments}\label{sec:experiments}
The experiment is conducted on each components of the proposed framework shown in Fig. \ref{fig:cnn+lstm} and \ref{fig:lstm} independently.  We follow the same experimental setup as indicated in \cite{Smedt2016} that used a leave-one subject-out cross-validation strategy. Based on this strategy, each proposed network is trained on 19 subjects and tested on the remaining subject, thus resulting in a 20-iteration cross validation. 

The depth-based CNN network is trained for 20 epochs using a min-batch size of 32 with the Adadelta optimizer \cite{Adadelta} with the default parameters of $lr=1.0$, $\rho=0.95$, and $\epsilon=1e^{-08}$.  The input of the network is based on the cropped hand images from the DHG-14/28 dataset re-sized to 227x227.  A low amount of epochs was chosen because the weights are designed to initialize the weights in the CNN+LSTM network therefore is not required to find the best optimal solution.

Similarly, the depth-based CNN-LSTM is using the Adadelta optimizer with default parameters with input image size of 227x227.  The network is trained with a mini-batch size of 16, timestep of 32, and in 100 epochs.  A timestep of 32 was selected because the number of key images for a gesture varies between 7 and 149 with an average of 34.59 key frames.

For the skeleton-based LSTM network, the Adam optimizer \cite{Adam} with default parameters of $lr=0.001$, $\beta_1=0.9$, $\beta_2=0.999$, and $\epsilon=1e^{-08}$ is used to train the network.  A timestep of 32 and a mini-batch size of 32 were selected becase it corresponds to the number of average key frames for each gesture.  Note that the input data is represented by the 2D coordinate points indicating the skeleton joints of a hand available from the DHG-14/28 dataset.

For the score-level and feature-level fusion networks, the Adadelta optimizer with default parameters is used.  The input data consists of sequences of depth images (227x227) and sequences of 2D skeleton joint locations (44x1).  The training is ran for 100 epochs with a timestep of 32 and a mini-batch size of 16.

\subsection{Datasets}
The dynamic hand gesture 14/28 (DHG-14/28) \cite{Smedt2016} was chosen as the database, and this is one of the few databases containing data collected using a depth camera (Intel RealSense F200) sensor. Both depth and skeleton information for various hand gestures is available.  In the DHG-14/28 \cite{Smedt2016} dataset, there are 20 unique individuals performing 5 iterations of 14 gestures using two types of finger configurations, thus forming 28 sets of gestures, to a total of 2800 sequences. The depth information is saved in the form of images with resolution of 480x640 in 16-bits. The skeleton information contains 22 joint locations of a hand described in both 2D and 3D coordinates saved in  44x1 and 66x1 vector format, respectively.

\begin{figure}[!ht]
\begin{center}
\begin{tabular}{cccc} 
\includegraphics[width=0.11\textwidth]{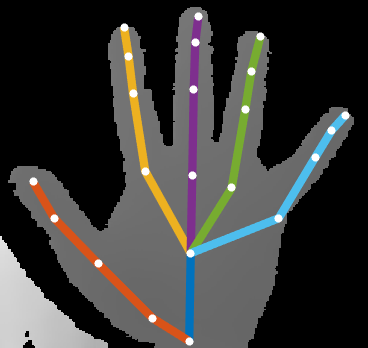} &
\includegraphics[width=0.094\textwidth]{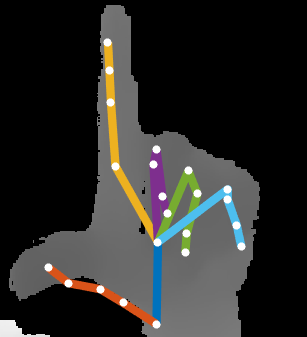} &
\includegraphics[width=0.076\textwidth]{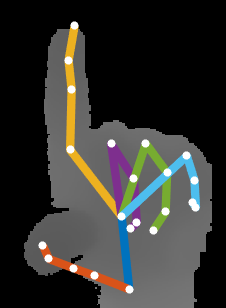} &
\includegraphics[width=0.106\textwidth]{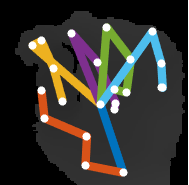} \\
$(a)$ & $(b)$ & $(c)$ & $(d)$
\end{tabular}
\caption{Illustrations of the grab gesture from the DHG-14/28 \cite{Smedt2016}.  $(a)$ Frame 1, $(b)$ Frame 40,  $(c)$ Frame 50,  $(d)$ Frame 60.
}\label{fig:activities}
 \end{center}
\end{figure}

For the DHG-14/28 dataset, each gesture is individually classified into two main categories: fine-grained and coarse-grained gestures.  Table \ref{tab:gestures} provides a list of all the gestures and the corresponding grain categories.
\begin{table}[!htb]
\centering
\caption{List of Gestures in the DHG-14/28 dataset}\label{tab:gestures}\vspace{-3mm}
\begin{scriptsize}
\begin{tabular}{|l|l|l|}
		\hline
		Gesture & Grain & Tag Name\\
		\hline
		\hline
		Grab & Fine & G\\
		Tap & Coarse & T\\
		Expand & Fine & E\\
		Pinch & Fine & P\\
		Rotation Clockwise & Fine & R-CW\\
		Rotation Counter-clockwise & Fine & R-CCW\\
		Swipe Right & Coarse & S-R\\
		Swipe Left & Coarse & S-L\\
		Swipe Up & Coarse & S-U\\
		Swipe Down & Coarse & S-D\\
		Swipe X & Coarse & S-X\\
		Swipe V & Coarse & S-V\\
		Swipe + & Coarse & S-+\\
		Shake & Coarse & Sh\\
		\hline
\end{tabular}
\end{scriptsize}
\end{table}

\subsection{Preprocessing}
The data inputs for the CNN+LSTM network are the depth images.  The depth images originally contain 16-bit information per pixel, and were normalized so that pixel value ranges from 0 to 1.  In addition, the hand images are cropped from the entire frame based on region of interest provided by the dataset.  Lastly, for each sequence, only the frames between the starting and ending motion of the gesture are used for recognition.

In  case of the LSTM network, only  2D skeleton points are used for processing.  For a selected gesture, every skeleton point in a sequence is normalized by subtracting every point by the palm location from the initial frame.  In addition, only the sequences marked between the start and the end of the gesture  are used for recognition.

\subsection{Experimental Results}
In order to compare the performance of the proposed method, we included the recognition rates of other methods that have also used the same database for experiments.  Table \ref{tab:14GestureRates} illustrates 14 unique gestures and the recognition rates of selected methods examined using the DHG-14/28 dataset. As mentioned in \cite{Chen2017}, it is not sufficient to provide only the averaged classification rates. Therefore for comparison, the best, worst, and the average classification rates for each gesture grain categories are provided in Table \ref{tab:14GestureRates}.  In addition, the results in Table \ref{tab:14GestureRates} are further examined based on the gesture's grain (fine, coarse, and/or both types of grain) as categorized previously in Table \ref{tab:gestures}.  

The conducted experimental study shows that both the proposed method of using the depth-based CNN+LSTM and the skeleton-based LSTM network demonstrate relatively similar performance.  Rows 4-6 in Table \ref{tab:14GestureRates} represent the performance of the proposed fusion networks.  FL-fusion-Concat shows the recognition rates achieved by the  feature-level fusion through concatenating the features extracted from the skeleton-based LSTM network and the depth-based CNN+LSTM network.  SL-fusion-Average reports the performance obtained by the score-level fusion through averaging the results of the soft-max layers of each of the skeleton and the depth-based networks.  SL-fusion-Maximum represents the  score-level fusion that predicts the output based on finding the maximum scores between the skeleton and the depth network.  

Of the three fusion networks, SL-fusion-Average performs the best, while FL-fusion-Concat performs the worst.  From Table \ref{tab:14GestureRates}, FL-fusion-Concat (row 4) performs worse than the default Skeleton LSTM network, which indicate that the process of fusion of depth and skeleton information at a feature level degrades the overall performance. It should be noted that even though the overall performance is reduced, the recognition rate for the depth-based fine-grained gesture is 3.60\% higher than the skeleton-based fine-grained gesture. SL-fusion-Average (row 5) and SL-fusion-Maximum (row 6) show similar performance with SL-fusion-Average performing 0.1\% better for both types of grained gestures. The performance of score-level fusion shows that this level of fusion results in better performance when compared to the independent skeleton and depth networks.  Even though skeleton network provides higher performance independently, the combined score between both networks yields the best performance because the depth provides information that may be lost in the process of skeleton joints extraction.

\begin{table*}[!htb]
\centering
\caption{Recognition Rates (\%) of the DHG-14 dataset}\label{tab:14GestureRates}\vspace{-3mm}
\begin{scriptsize}
\begin{tabular}{|l|c|c|c|c|c|c|c|c|c|}
		\hline
		\multicolumn{1}{|c|}{\multirow{2}{*}{Method}}& \multicolumn{3}{|c|}{Fine}& \multicolumn{3}{|c|}{Coarse}& \multicolumn{3}{|c|}{Both}\\
		\cline{2-10}
		& Best & Worst & Avg $\pm$ Std & Best & Worst & Avg $\pm$ Std & Best & Worst & Avg $\pm$ Std \\
		\hline
		\hline
		Depth CNN & 52.89& 24.83& 37.05$ \pm$ 7.89& 38.92& 20.61& 29.57$ \pm $4.51& 40.19& 25.05&	32.24 $\pm$ 4.64 \\
		\hline
		Depth CNN+LSTM &90.00 &52.00 &73.50 $\pm$ 10.93 &92.22 &58.89 & 77.06 $\pm$ 8.73&85.00 & 58.57&	75.79 $\pm$ 7.23\\
		\hline
		Skeleton LSTM	& 82.00& 54.00&69.90 $\pm$ 9.91 &96.67 &76.67 &89.00 $\pm$ 5.40 & 91.43& 71.43& 82.18 $\pm$ 5.32\\
		\hline
		FL-fusion-Concat	& 90.00& 48.00&72.90 $\pm$ 10.30 &\bf{98.89} &78.89 &86.83 $\pm$ 4.68 & 87.86& 67.86& 81.86 $\pm$ 5.38\\
		\hline
		SL-fusion-Average & 92.00& 52.00&76.00 $\pm$ 10.51 &97.78 &\bf{81.11} &90.72 $\pm$ 4.64 & \bf{95.00}& \bf{72.86}& \bf{85.46 $\pm$ 5.16}\\
		\hline
		SL-fusion-Maximum & \bf{94.00}& 54.00&75.30 $\pm$ 10.89 &\bf{98.89}&78.89 &\bf{90.94 $\pm$ 4.36}& 91.43& 71.43& 85.36 $\pm$5.06\\
		\hline
		Skeleton \cite{Chen2017}	& 86.00& 42.00&61.20 $\pm$ 12.37 &97.78 &74.44 &86.44$\pm$ 7.94 & 93.57& 67.86& 77.43 $\pm$ 6.82\\
		\hline
		Motion Feature \cite{Chen2017}	& 84.00& 46.00&71.50 $\pm$ 11.44 &96.67 &64.44 &81.94 $\pm$ 8.17 & 90.00& 58.57& 78.21 $\pm$ 7.49\\
		\hline
		Skeleton + Motion Feature \cite{Chen2017}	& 90.00& \bf{56.00}& \bf{76.90 $\pm$ 9.19} &97.78 &72.22 &89.00 $\pm$ 7.55 & 94.29& 67.86& 84.68 $\pm$ 6.67\\
		\hline
		3D  Skeleton CNN+LSTM \cite{Nunez2018}	& N/A& N/A&\bf{78.00} & N/A&N/A &89.83 &N/A &N/A & \bf{85.61}\\
		\hline
		Skeleton-based,  De Smedt \cite{Smedt2016}	&N/A & N/A&73.60 &N/A &N/A &88.33 & N/A& N/A& 83.07\\
		\hline
		Depth-based, De Smedt \cite{Smedt2016}	& N/A& N/A&66.90& N/A& N/A&85.94 &N/A & N/A& 79.14\\
		\hline
\end{tabular}
\end{scriptsize}
\end{table*}

\begin{figure*}[!ht]
\begin{center}
\includegraphics[interpolate,width=\textwidth]{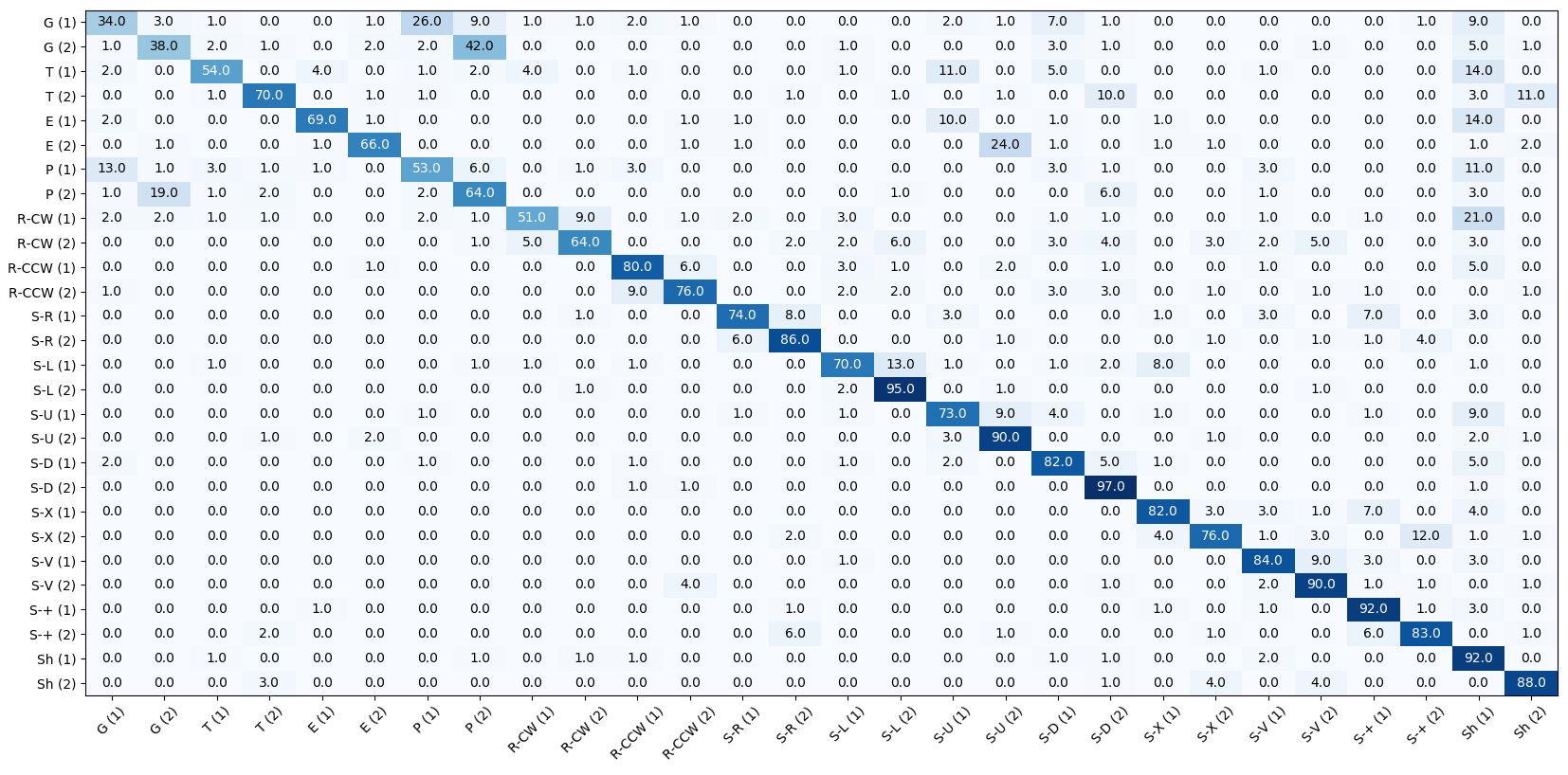}
\caption{The confusion matrix showing accuracies of gesture recognition when using the score level fusion (average) of both skeleton and depth information on 28 gestures.}
\label{fig:28gcm}
 \end{center}
\end{figure*}

\begin{figure}[!ht]
\begin{center}
\includegraphics[interpolate,width=0.44\textwidth]{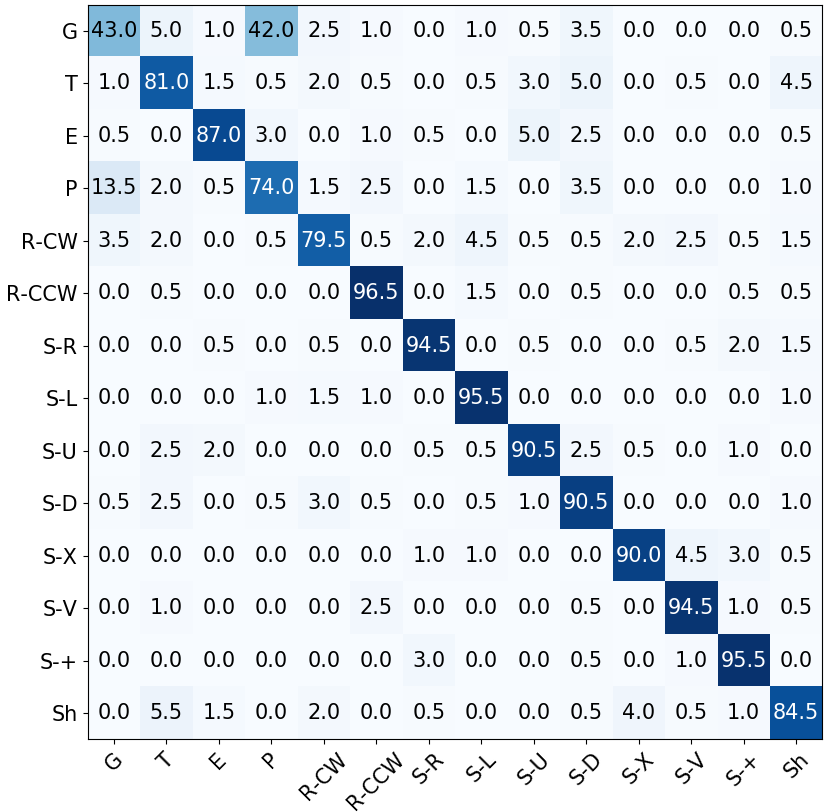} 
\caption{The confusion matrix showing accuracies of gesture recognition when using the score level fusion (average) of both skeleton and depth information on 14 gestures.}
\label{fig:smfusion-ave}
 \end{center}
\end{figure}

To further examine the performance of the gesture recognition in terms of the averaged classification rates, a confusion matrix was created (Fig. \ref{fig:smfusion-ave} and \ref{fig:28gcm}). It illustrates the accuracy of the classification performed by the proposed method when using score-level fusion of skeleton and depth information for 14 and 28 gestures, respectively. For each figure, the x-axis represents the network's prediction whereas the y-axis represents the true gesture.  For example, the gesture R-CW (5th row) indicates that the network is able to predict 79.5\% of all the R-CW gestures  correctly, while misidentifying the other 20.5\% as other gestures.  Fig. \ref{fig:smfusion-ave} shows that the biggest failure of the fusion network occurs at the prediction of the grabbing gesture (1st row) which is misidentified 67.0\% of the time of which 42.0\% is classified as the pinching gesture.  This specific grabbing gesture has been noted by \cite{Chen2017} and \cite{Smedt2016} to be difficult to distinguish from the pinching motion.  Similar observation can also be seen in Fig. \ref{fig:28gcm}.

Based on the gesture recognition rates from Fig. \ref{fig:smfusion-ave} and \ref{fig:28gcm}, the Loss of Accuracy when Removing the Finger Differentiation (suggested in \cite{Smedt2016}) is calculated to be 0.07286 which is seven times higher than 0.0114 obtained in \cite{Smedt2016}.  On average, the fusion provided a recognition rate of 85.46\% and 74.19\% for 14 and 28 gestures, respectively.  The transition between 14 to 28 gestures results in a 11.27\% decrease of which 7.29\% comes from intra-gestures confusion.  This  indicates that the fusion network is not yet optimized for the 28 gestures, and by using the network weights pre-trained on 14 gestures, the overall network produces greater error.

\section{Conclusions}\label{sec:conclusions}

This paper contributes to the study of ambient, sensor-based monitoring of human activity and gesture recognition. Specifically, it focuses on the task for dynamic hand gesture recognition, using both skeleton and depth data acquired by depth-RGB sensors, and the powerful deep learning techniques. The results of the  conducted experimental study are summarized as follows:
\begin{enumerate}
	\item Usage of depth data, in addition to visual spectrum, RGB data, has potential to greatly improve the performance of hand gesture recognition algorithms that rely on temporal patterns.  While usage of only depth data with CNN leads to the average recognition rate of 32.24\%. However, utilizing the CNN for feature extraction, and passing those features to the RNN, the overall performance is shown to increase up to 75.79\%.  This results indicate that CNN alone is insufficient in extracting enough information from a singular depth image to correctly predict the desired gesture.  By combining the CNN with the RNN, the new network is able to recognize patterns from the features extracted from a CNN throughout multiple frames.  Another aspect influencing the performance is the chosen timestep parameter which depending on the input data may be severely truncated when given long sequences or padded with blank information for short sequences.
	\item The performance of using skeleton data has shown a recognition rate of 82.18\% which is higher compared to the depth-based networks.  However, when examining the recognition rates for each grain gesture, the depth-based approach produces a higher recognition rate for the fine-grained gesture.  Based on this observation, the fusion of both depth-based and skeleton-based networks should yield a higher overall recognition rate.  Experiments show that by using a score-level fusion, a recognition rate of 85.46\% is achieved, which is higher than the rates of 82.18\% and 76.50\% for the independent skeleton and the depth based approaches, respectively.  In addition to an overall higher recognition rates, the rates for each type of grain gesture also increased.  For the fine grained gestures, while the rate for separately obtained depth-based  and skeleton-based  approach are 73.50\% and 67.20\%, respectively, the combined rate is 76.00\%. For the coarse grained gestures, the depth-based (78.17\%) and skeleton-based (89.00\%), when combined, show the rate of 90.72\%. 
	\item The applied fusion allowed to achieve very good performance for the dataset consisting of 14 gestures.  When the same approach is applied to the 28 gesture version, the performance degrades to 74.19\% which is an $\approx$11\% decrease.  The main reason for this decrease is that the networks' weights are optimized for recognizing 14 gestures.  Therefore, a better performance can achieved by training each network independently for recognizing 28 gestures prior to fusion. 
\end{enumerate}
As a future work, we anticipate to expand the proposed framework toward human activity recognition which would be applied to various assisting living, healthcare and human-machine interaction scenarios.  

\section*{Acknowledgments}
\begin{small}
This project was partially supported by Natural Sciences and Engineering Research Council of Canada (NSERC) through Discovery Grant ``Biometric Intelligent Interfaces''; the Province of Alberta Queen Elizabeth II Scholarship and the University of Calgary W21C (VPR's grant ``Innovations in Home Health Care to Support an Aging Population'').
\end{small}

{\small
\bibliographystyle{IEEEtran}
\bibliography{gest}
}

\end{document}